\documentclass{article}

\usepackage[preprint, nonatbib]{neurips_2024}
\usepackage[utf8]{inputenc} 
\usepackage[T1]{fontenc}    
\usepackage{url}            
\usepackage{booktabs}       
\usepackage{nicefrac}       
\usepackage{microtype}      
\usepackage{biblatex} 
\usepackage{array}
\usepackage[caption=false,font=normalsize,labelfont=sf,textfont=sf]{subfig}
\usepackage{textcomp}
\usepackage{stfloats}
\usepackage{lscape}
\usepackage{multirow}
\usepackage{longtable}
\usepackage{caption}
\usepackage[hidelinks]{hyperref}
\usepackage[dvipsnames]{xcolor}
\usepackage{centernot,cancel}
\usepackage{lipsum}
\usepackage{soul,balance}
\usepackage{pifont}
\usepackage{amsmath,amsfonts,amssymb,amsthm,amstext}
\usepackage{verbatim}
\usepackage{graphicx}
\usepackage{siunitx}
\usepackage{algorithm}%
\usepackage{algpseudocode}
\usepackage{etoolbox} 
\usepackage{svg}
\usepackage{epstopdf}
\usepackage{import}

\title{DQS: A Low-Budget Query Strategy for Enhancing Unsupervised Data-driven Anomaly Detection Approaches}

\author{%
  Lucas Correia\\
  Leiden University \\
  Leiden \\
  The Netherlands \\
  \texttt{l.ferreira.correia@liacs.leidenuniv.nl} \\
  \And
  Jan-Christoph Goos \\
  Mercedes-Benz AG\\
  Stuttgart\\
  Germany \\
  \AND
  Thomas Bäck \\
  Leiden University \\
  Leiden \\
  The Netherlands \\
  \And
  Anna V. Kononova \\
  Leiden University \\
  Leiden \\
  The Netherlands \\
\texttt{a.kononova@liacs.leidenuniv.nl} \\
}

\begin{document}

\maketitle

\begin{abstract}
Truly unsupervised approaches for time series anomaly detection are rare in the literature.
Those that exist suffer from a poorly set threshold, which hampers detection performance, while others, despite claiming to be unsupervised, need to be calibrated using a labelled data subset, which is often not available in the real world.
This work integrates active learning with an existing unsupervised anomaly detection method by selectively querying the labels of multivariate time series, which are then used to refine the threshold selection process.
To achieve this, we introduce a novel query strategy called the dissimilarity-based query strategy (DQS).
DQS aims to maximise the diversity of queried samples by evaluating the similarity between anomaly scores using dynamic time warping.
We assess the detection performance of DQS and explore the impact of mislabelling, a topic that is underexplored in the literature.
Our findings indicate that DQS performs best in small-budget scenarios, though the other methods appear to be more robust when faced with mislabelling.
Regardless, all query strategies outperform the unsupervised threshold even in the presence of mislabelling.
Thus, whenever it is feasible to query an oracle, employing an active learning-based threshold is recommended.
\end{abstract}

\section{Introduction}
With the increasing digitisation of industrial processes, the amount of recorded data continues to grow. 
Ensuring that this data accurately represents the respective process is crucial, as incomplete or contaminated data can adversely affect downstream tasks such as modelling and optimisation.
For tasks involving system behaviour modelling, data that deviates from the norm is undesirable and is referred to as \emph{anomalous} behaviour.
The antonym most commonly used for \emph{anomalous} is \emph{normal}, though to avoid confusion with a Gaussian distribution, \emph{nominal} is used to refer to anomaly-free data henceforth.
The format of recorded data varies across applications and domains, with time series being a common representation in applications such as cardiology~\cite{moody_impact_2001}, server performance monitoring~\cite{su_robust_2019}, water management systems~\cite{mathur_swat_2016, ahmed_wadi_2017}.

The field of unsupervised time series anomaly detection has seen many contributions being published recently, though many of the approaches are actually either semi-supervised~\cite{chandola_anomaly_2012} (they use nominal-only historic data for model training~\cite{wang_variational_2022, park_multimodal_2018, niu_lstm-based_2020, zhang_federated_2021, choi_multivariate_2022}) or are actually supervised (they use some amount of labelled data for hyperparameter tuning~\cite{tafazoli_matrix_2023, fahrmann_lightweight_2022, li_mad-gan_2019, zhang_acvae_2024, chen_iot-gan_2022} and threshold setting~\cite{tuli_tranad_2022, chen_learning_2021, su_robust_2019, chen_unsupervised_2020, von_schleinitz_vasp_2021}).
Truly unsupervised approaches exist~\cite{correia_tevae_2024}, but especially the threshold setting poses a challenge.
Correia et al.~\cite{correia_tevae_2024} illustrate the disparity in detection performance between the temporal variational autoencoder (TeVAE) approach in its unsupervised mode and TeVAE if it were combined with the best possible threshold, indicating a large potential remaining untapped due to the threshold setting.

One way to reach said untapped potential would be to employ active learning, where obtaining labelled samples from an oracle may allow for a threshold choice that leads to a higher detection performance.
This would turn a purely unsupervised approach into something both unsupervised and supervised, or arguably just supervised, though unlike the cited literature above, it would not assume labelled data for threshold setting is magically available at no cost.

In this work, we propose a novel query strategy which is non-parametric, i.e.\ it does not need to be parameterised ahead of use.
It works by estimating the similarity of samples by calculating their respective dynamic time warping distance to maximise the diversity of samples queried.

Once the query set is labelled by an oracle, it can be used to search for a threshold that corresponds to better detection performance.
Additionally, literature in the field of active learning in time series anomaly detection has always assumed a perfect oracle, i.e.\ all labels provided are correct.
For a more detailed overview, please refer to Section~\ref{sec:related_work}.
However, this is an idealised assumption of the real world, and hence we also provide a novel testing methodology that accounts for the probability of oracle mislabelling.

This paper is structured as follows.
In Section~\ref{sec:background}, a brief background is provided on active learning, as well as the theory behind computing the dynamic time warping distance.
Additionally, the mathematical problem description is outlined.
Following that, in Section~\ref{sec:related_work}, literature related to this paper is presented, including an analysis on the research gap identified.
Then, in Section~\ref{sec:proposed_approach}, we propose the dissimilarity-based query strategy (DQS), which aims to fill the research gap identified in Section~\ref{sec:related_work}.
In Section~\ref{sec:experiments} the experiment procedure is shown along with the respective results.
Lastly, in Section~\ref{sec:conclusion}, conclusions are made and an outlook on future research is provided.

The source code corresponding to this paper can be found in the form of a repository under \href{https://github.com/lcs-crr/DQS}{\texttt{github.com/lcs-crr/DQS}}, and the PATH dataset can be downloaded from \href{https://zenodo.org/records/13255120}{\texttt{zenodo.org/records/13255120}}~\cite{correia_dataset_2024}.

\section{Background} \label{sec:background}
\subsection{Active Learning}
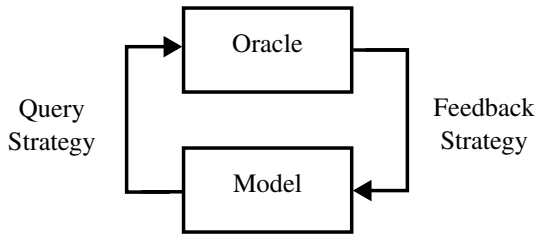
\begin{figure}[h!]
    \centering
    \def\svgwidth{0.5\textwidth}
\begingroup%
  \makeatletter%
  \providecommand\color[2][]{%
    \errmessage{(Inkscape) Color is used for the text in Inkscape, but the package 'color.sty' is not loaded}%
    \renewcommand\color[2][]{}%
  }%
  \providecommand\transparent[1]{%
    \errmessage{(Inkscape) Transparency is used (non-zero) for the text in Inkscape, but the package 'transparent.sty' is not loaded}%
    \renewcommand\transparent[1]{}%
  }%
  \providecommand\rotatebox[2]{#2}%
  \newcommand*\fsize{\dimexpr\f@size pt\relax}%
  \newcommand*\lineheight[1]{\fontsize{\fsize}{#1\fsize}\selectfont}%
  \ifx\svgwidth\undefined%
    \setlength{\unitlength}{2250bp}%
    \ifx\svgscale\undefined%
      \relax%
    \else%
      \setlength{\unitlength}{\unitlength * \real{\svgscale}}%
    \fi%
  \else%
    \setlength{\unitlength}{\svgwidth}%
  \fi%
  \global\let\svgwidth\undefined%
  \global\let\svgscale\undefined%
  \makeatother%
  \begin{picture}(1,0.43333293)%
    \lineheight{1}%
    \setlength\tabcolsep{0pt}%
    \put(0,0){\includegraphics[width=\unitlength,page=1]{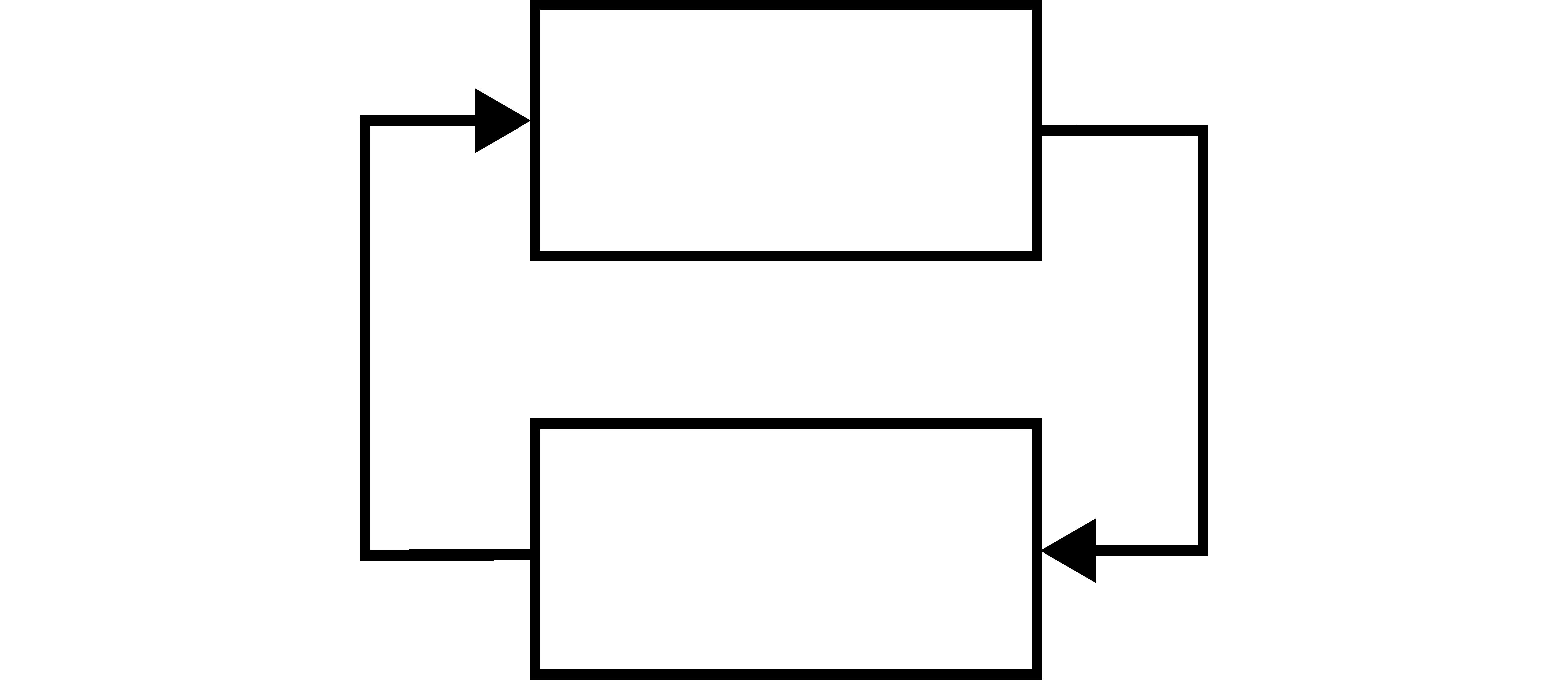}}%
    \put(0.50101857,0.34589803){\makebox(0,0)[t]{\lineheight{1.25}\smash{\begin{tabular}[t]{c}Oracle\end{tabular}}}}%
    \put(0.50121388,0.0792318){\makebox(0,0)[t]{\lineheight{1.25}\smash{\begin{tabular}[t]{c}Model\end{tabular}}}}%
    \put(0.9134206,0.22458901){\makebox(0,0)[t]{\lineheight{1.25}\smash{\begin{tabular}[t]{c}Feedback\\Strategy\end{tabular}}}}%
    \put(0.09051152,0.22245918){\makebox(0,0)[t]{\lineheight{1.25}\smash{\begin{tabular}[t]{c}Query\\Strategy\end{tabular}}}}%
  \end{picture}%
\endgroup%

    \caption{Interaction between the oracle and a model-based anomaly detector.}
    \label{fig:strategies}
\end{figure}

The oldest literature in the field of active learning deals mostly with data augmentation for supervised classification algorithms~\cite{settles_active_2012}, though lately, it has also found application in anomaly detection.
In the context of model-based anomaly detection, active learning typically involves two components in a loop, as shown in Figure~\ref{fig:strategies}.

The first component is the oracle, also referred to as a user, human annotator, or domain expert, whose task is to label previously unlabelled samples as nominal or anomalous, so they can be used for other tasks.
The second component is the anomaly detection model, which is trained in10 an unsupervised manner, i.e.\ on unlabelled data, or in a semi-supervised manner, i.e.\ on nominal-only data.
The samples labelled by the oracle are used to calibrate the model to improve detection performance.
The oracle and the model interact with each other using the query strategy and the feedback strategy.

The query strategy is defined as the procedure used to choose the samples to be sent to the oracle~\cite{zhou_boundary-driven_2024}.
The goal of the query strategy is to select the samples from a candidate pool that provide the most useful information for the feedback strategy.
Labelling is expensive, and hence the query strategy is often constrained by a parameter called the query budget, which represents the number of samples an oracle is willing to label at a time.

The feedback strategy is defined as the method of applying the knowledge in form of the labelled samples.
In the case of data augmentation supervised classification algorithms, the feedback strategy is straight-forward, however, in the unsupervised anomaly detection, it plays a larger role due to the inherent incompatibility of unsupervised methods with labelled data.

\subsection{Dynamic Time Warping}
Originally proposed by Vintsyuk~\cite{vintsyuk_speech_1972} and Sakoe and Chiba~\cite{sakoe_dynamic_1978}, dynamic time warping (DTW) is used to calculate the optimal alignment of two time series that may be of different lengths and out of sync w.r.t.\ each other.

To calculate the DTW distance $\delta$, consider two univariate sequences $\mathbf{x}$ and $\mathbf{y}$, where $\mathbf{x} \in \mathbb{R}^{T_\mathbf{x}}$, $\mathbf{y} \in \mathbb{R}^{T_\mathbf{y}}$, and $u$, $v$ represent arbitrary indices within the respective sequences, such that $\mathbf{x} = [x_1, ..., x_u, ..., x_{T_\mathbf{x}}]$ and $\mathbf{y} = [y_1, ..., y_v, ..., y_{T_\mathbf{y}}]$.
The local cost matrix $\mathbf{C}$ represents the cost of aligning each time step in $\mathbf{x}$ with each time step in $\mathbf{y}$. 
The implementation of DTW used in this work~\cite{meert_dtaidistance_2020} uses the Euclidean norm for the computation of the local cost.
Hence, the local cost between $\mathbf{x}$ at index $u$ and $\mathbf{y}$ at index $v$, as shown in Equation~\ref{eq:local_cost}.
\begin{equation}
    \mathbf{C}_{u,v} = (x_u-y_v)^2
    \label{eq:local_cost}
\end{equation}

The optimal path through $\mathbf{C}$ represents the path where the cumulative cost is lowest and can found using dynamic programming.
In cumulative cost matrix $\mathbf{D}$ each index denotes the cumulative cost of arriving at the respective index; it is calculated as shown in Equation~\ref{eq:cumulative_cost}.
\begin{equation}
\mathbf{D}_{u, v} = \mathbf{C}_{u, v} + \min
    \begin{cases}
        \mathbf{D}_{u-1, 1} & \text{for   } u > 1 \\
        \mathbf{D}_{1, v-1} & \text{for   } v > 1 \\
        \mathbf{D}_{u-1, v-1} & \text{for   } u,v > 1
    \end{cases} 
\label{eq:cumulative_cost}
\end{equation}

Finally, the DTW distance $\delta$ equals the square root of the cumulative cost of the optimal path, as shown in Equation~\ref{eq:dtw_distance}.
\begin{equation}
    \delta = \sqrt{\mathbf{D}_{{T_\mathbf{x}}, {T_\mathbf{y}}}}
    \label{eq:dtw_distance}
\end{equation}

\subsection{Problem Description}
Consider an anomaly detection problem involving some sort of dynamic process running for several days that yields a number of multivariate time series that are sequential w.r.t.\ each other but are not necessarily temporally contiguous, i.e.\ a discrete-sequence anomaly detection problem~\cite{correia_online_2024}.
Note that, in this work, \emph{sequence} and \emph{time series} are used synonymously.
During the dynamic process, each recorded time series $\mathcal{S}_{n}$ is added to the candidate set $\mathcal{C}^{\text{ts}} = \{\mathcal{S}_{1}, ..., \mathcal{S}_{n}, ..., \mathcal{S}_{N}\}$ containing all recorded sequences up to the present, where $N$ denotes the number of sequences recorded up to the present and $\mathcal{S}_{n} \in \mathbb{R}^{T_{n} \times d_\mathcal{D}}$, where $T_{n}$ denotes the number of time steps in $\mathcal{S}_{n}$, and $d_\mathcal{D}$ the dimensionality of the data set.
Now, consider a model $\mathcal{M}$ that has been used to check the data for anomalous behaviour and yields a univariate anomaly score $\mathbf{s}_{n} = \mathcal{M}(\mathcal{S}_{n})$, where $\mathbf{s}_{n} \in \mathbb{R}^{T_{n}}$.
The anomaly score counterpart to the candidate set $\mathcal{C}^{\text{ts}}$ is denoted as the candidate score set $\mathcal{C}^{\text{as}}$ and is structured similarly, such that $\mathcal{C}^{\text{as}} = \{\mathbf{s}_{1}, ..., \mathbf{s}_{n}, ..., \mathbf{s}_{N}\}$.

In the context of active learning, a sequence is treated as a sample, where a subset of sequences of $\mathcal{C}^{\text{ts}}$ is chosen to be sent to the oracle.
Once the query budget $B$ is used up, i.e.\ $B$ time series are chosen, they are moved to a set called the query set $\mathcal{Q}^{\text{ts}}$ which contains all time series selected for querying the oracle up to the present. 
The anomaly score counterpart to the query set $\mathcal{Q}^{\text{ts}}$ is denoted as the query score set $\mathcal{Q}^{\text{as}}$ and is structured similarly. 
Once the query score set is established, it can be used to improve the anomaly detection performance using the feedback strategy.

\section{Related Work} \label{sec:related_work}
The wide majority of model-based approaches found in the literature use some sort of anomaly score, which indicates how anomalous each time step within a test subset sequence is.
To classify time steps in the anomaly score as nominal or anomalous, a threshold needs to be set, however, many publications in unsupervised anomaly detection circumvent the issue by using labelled data, technically making the respective approaches supervised.
The labelled data is either used to set a threshold which satisfies a condition, like achieving the maximum possible $F_1$ score or some other parametric threshold setting method is proposed, like the peaks over threshold (POT) method~\cite{su_robust_2019}, which has found wide application throughout the literature~\cite{fan_luad_2023, tuli_tranad_2022, zhang_unsupervised_2021, yu_dtaad_2024}.
Hundman et al.~\cite{hundman_detecting_2018} introduce the non-parametric dynamic threshold, but unfortunately, despite \emph{non-parametric} being in the name, it is not truly non-parametric, as the threshold can only be set by setting a hyperparameter $\mathbf{z}$ which is experimentally set, and therefore requires labelled data.
In addition to that, this method cannot be applied to discrete-sequence anomaly detection, as it assumes a predicted anomalous time steps or ground-truth anomalous time steps within a given test sequence. 
When neither is given, like in the case of a correctly identified nominal time series, the threshold becomes undefined.

In contrast, few truly unsupervised threshold setting methods are proposed in the literature.
The most naive method is using the training subset $\mathcal{D}^\text{train}$ or validation subset $\mathcal{D}^\text{val}$ to set a threshold, as implemented by, for example, Correia et al.~\cite{correia_ma-vae_2023, correia_tevae_2024} or Zhang et al.~\cite{zhang_federated_2021}. 
Throughout this work, the threshold obtained using this method is referred to as \emph{unsupervised threshold} $\tau_\text{us}$. 
This method is motivated by the goal of achieving high precision, since the maximum training or validation subset error is likely to be at or higher than the highest error resulting from nominal sequences in the test subset.
Its simplicity is at the same time also a weakness, as a single scalar value is used to separate nominal from anomalous samples in an anomaly score, which are functions of time.
Additionally, in an unsupervised setting there may be anomalous sequences in the training and validation subsets, which lead to a higher maximum anomaly score and therefore a threshold that is set higher than optimal, as can be observed in, for example, Correia et al.~\cite{correia_dataset_2024-1}.

Generally, active learning is rarely used to adjust thresholds in model-based time series anomaly detection approaches.
Literature dealing with further training the anomaly detection model exists, but such methods are not generically applicable, since they modify loss functions to penalise wrong classifications~\cite{tang_deep_2020, huang_semi-supervised_2022, wang_active-mtsad_2022, li_situation-aware_2022} or utilise an architecture specific to active learning which also requires labelled data from the beginning~\cite{yu_amad_2024}.

Lundström et al.~\cite{lundstrom_interactive_2023} propose a threshold setting method for continuous-sequence problems.
First, it initialises a threshold based on a variation of the non-parametric dynamic threshold~\cite{hundman_detecting_2018}, then anomalous sub-sequences are clustered according to four pre-defined attributes of temporally adjacent sub-sequences.
According to Lundström et al.~\cite{lundstrom_interactive_2023}, the mathematical definition of \emph{temporally adjacent} implies that two sub-sequences can be temporally adjacent if they occur after one another, regardless of whether there is another sub-sequence between them, though we interpret their explanation as referring to two sub-sequences following one another \emph{directly}, i.e.\ there is no other sub-sequence between them, since they would not otherwise be \emph{adjacent}.
The first attribute, $A_1$ denotes the temporal distance between the last time step of a predicted anomalous sub-sequence and the first time step of the following predicted anomalous sub-sequence.
The second and third attributes, $A_2$ and $A_3$, denote the absolute difference between the respective maximums of two temporally adjacent predicted anomalous sub-sequences.
The difference between the second and third attributes is that the former looks for the maximum value in each channel of the input time series, whereas the latter looks for the maximum value in each channel of the anomaly score.
The fourth attribute $A_4$ is a binary vector, which assumes a $1$ if both maximum anomaly scores in each channel of two temporally adjacent subsequences exceed the current threshold.
As is evident, all but the first attribute are obtained feature-wise, i.e.\ they $A_2, A_3, A_4 \in \mathbb{R}^{d_\mathcal{D}}$.
Following the calculation of attributes, Lundström et al.~\cite{lundstrom_interactive_2023} aggregate them into several logical outputs, which assume a binary value depending on whether the respective attribute exceeds the set threshold for that attribute.
The logical outputs are then combined to decide whether two temporally adjacent predicted anomalous sub-sequences should be clustered together.
Ironically, this threshold setting method requires pre-defined thresholds to work, and therefore it needs to be parametrised.
Additionally, it cannot be applied to discrete-sequence problems since, to obtain attribute $A_1$, predicted anomalous sub-sequences need to be within the same sequence, which may not always be the case.

There is a clear need for methods which allow for the integration of active learning into the threshold choice for discrete-sequence anomaly detection problems.
Additionally, all the above-mentioned literature assumes perfect labelling, however, as pointed out by  Settles~\cite{settles_active_2012}, labels from human experts are not always reliable as some samples are difficult to label and humans can be distracted or fatigued over time.
Therefore, this work investigates the impact of different rates of mislabelling on the anomaly detection performance, which has not been undertaken previously in the literature.

\section{Proposed Approach} \label{sec:proposed_approach}
In this paper, a novel query strategy is proposed, named the dissimilarity-based query strategy (DQS).
Generally, query strategies work by letting an oracle label $B$ times each round, i.e.\ the number of samples allowed to be queried in a given round $q$. 
While it is possible to choose $B$ sequences randomly every round, there is little control over the diversity of the queried sequences.
In this work, the diversity of sequences is maximised by actively choosing samples most dissimilar to the previously queried ones, by inspecting the DTW distance between the samples in question, i.e.\ the anomaly scores of different time series, where we interpret a high distance as poor alignment of features within the anomaly scores and, therefore, a high level of dissimilarity.
Unlike other metrics, like the mean-squared-error, dynamic time-warping distance does not require two sequences to be the same length, which makes it ideal for an application with variable-length sequences.
Calculating the dissimilarity between multivariate time series in the same way is possible, however, it is much more computationally intensive. 

As mentioned, the unsupervised threshold $\tau_\text{us}$ discussed in Section~\ref{sec:related_work} is based on the data in the validation subset, and hence the validation subset is henceforth used as the candidate set.
This is because the sequences in the training subset have already been used to fit the model and therefore will be reconstructed with a lower anomaly score, skewing results.
The exact procedure for DQS is as follows and is also depicted as pseudocode in Algorithm~\ref{alg:ds_query_mech}.
\begin{algorithm}[h!]
\caption{DS-based query strategy at an arbitrary round $q$}\label{alg:ds_query_mech}
\begin{algorithmic}[1]
\Require $\text{Candidate score set  } \mathcal{C}^{\text{as}}, \text{Query score set  } \mathcal{Q}^{\text{as}}, \text{Budget  } B$ 
\For{$b=1 \to B$}
    \If{$\vert\mathcal{Q}^{\text{as}}\vert=0$}
        \State $r \sim \mathcal{U}(1, \vert \mathcal{C}^\text{as} \vert)$ \Comment{Random index for initialisation}
    \Else
        \For{$n = 1 \to \vert \mathcal{C}^\text{as} \vert$}
            \For{$m=1 \to \vert \mathcal{Q}^\text{as} \vert$}
                \State $\mathbf{D}[n, m] = DTW(\mathcal{C}^\text{as}[n], \mathcal{Q}^\text{as}[m])$
            \EndFor
        \EndFor
        \State $o = \arg\min(\mathbf{D})$ \Comment{Index most similar to any sample in $\mathcal{Q}^\text{as}$}
        \For{$n=1 \to \vert \mathcal{C}^\text{as} \vert$}
            \State $\mathbf{E}[n] = DTW(\mathcal{C}^\text{as}[n], \mathcal{C}^\text{as}[o])$
        \EndFor
        \State $r = \arg\max(\mathbf{E})$ \Comment{Index most dissimilar to $\mathcal{C}^\text{as}[o]$}
    \EndIf
    \State $\mathcal{Q}^{\text{as}} \gets \mathcal{C}^\text{as}[r]$  \Comment{Move $\mathcal{C}^\text{as}[r]$ to $\mathcal{Q}^{\text{as}}$}
\EndFor
\State \textbf{return} $\text{Query Score Set  } \mathcal{Q}^{\text{as}}$
\end{algorithmic}
\end{algorithm}

For the first query in the first round, i.e.\ $b=q=1$ and $\vert\mathcal{Q}^{\text{as}} \vert=0$, DQS is initialised by choosing a random anomaly score with index $r$ from the candidate score set $\mathcal{C}^{\text{as}}$, as shown in line $3$ in Algorithm~\ref{alg:ds_query_mech}.
It is then moved to the query score set $\mathcal{Q}^{\text{as}}$, such that $\vert \mathcal{Q}^{\text{as}}\vert=1$, as shown in line $16$.
Once the querying is initialised, for the next query, i.e.\ $b=2$, the anomaly score in $\mathcal{C}^{\text{as}}$ most similar to any anomaly score within $\mathcal{Q}^{\text{as}}$ is highlighted; its index is denoted as $o$, as shown in lines $5$ to $10$.
Following that, the anomaly score in $\mathcal{C}^{\text{as}}$ most dissimilar to $\mathcal{C}^{\text{as}}[o]$, denoted by index $r$, is moved to $\mathcal{Q}^{\text{as}}$, such that $\vert \mathcal{Q}^{\text{as}}\vert=2$, as shown in lines $11$ to $14$.
This is repeated until $B$ sequences have been moved to $\mathcal{Q}^{\text{as}}$, such that, at the end of round $1$, $\vert\mathcal{Q}^{\text{as}}\vert = B$.
For round $2$, no initialisation is required, and hence the procedure is the same as round $1$ for $b>1$.
Therefore, after round $2$, $B$ anomaly scores are moved to the query score set $\vert\mathcal{Q}^{\text{as}}\vert = 2 \cdot B$.
The following rounds after that follow the same pattern.
Clearly, the query set and the candidate set both grow with every round, complicating computation as time goes on.

Once a round is over and $B$ new samples have been added to the query score set $\mathcal{Q}^{\text{as}}$, the threshold $\tau$ is grid-searched based on the labels provided by the oracle, though future work could investigate the feasibility of more efficient search methods.

This method is generically applicable to discrete-sequence problems that rely on a threshold as a decision boundary between nominal and anomalous sequences.

\section{Experiments}\label{sec:experiments}
\subsection{Impact of Mislabelling Probability}
While it is assumed the oracle in this work is a domain expert, they can still make mistakes.
In literature, the effect of mislabelling has not been investigated as far the authors are aware, therefore, for a fixed budget size of $B=10$ queries, a mislabelling probability of $p_m=0.1$, $p_m=0.2$,  $p_m=0.3$ is experimented with. 
This is done by sampling from a Bernoulli distribution parameterised with the above-mentioned mislabelling probabilities and flipping ground-truth labels accordingly.
The choice of the Bernoulli distribution is based on several reasons.
For one, unlike a Binomial distribution, which is a function of the number of trials, the Bernoulli distribution is of single-trial nature. 
Additionally, the Bernoulli distribution is discrete and binary, sampling it yields a boolean, corresponding to mislabelled or not mislabelled.

\subsection{Impact of Query Budgets}
A high query budget $B$ is associated with a higher labelling cost, and hence it is desirable to use a query strategy that maximises the $F_1$ score improvement with a limited number of queries.
To investigate the impact the query budget has on the anomaly detection results, three different budget sizes are used: $B=1$, $B=5$, and $B=10$. 
In this experiment, it is assumed that all queries are correctly labelled.

\subsection{Benchmarking Considerations}
This work performs all experiments on the PATH data set~\cite{correia_dataset_2024}, a discrete-sequence data set that provides a large number of diverse of multivariate time series and non-trivial anomalies.
The dataset represents the dynamic behaviour of an automotive powertrain at different states, which is generated using state-of-the-art simulation tools to exhibit real-world-like complexity.
Other publicly available datasets exist, though none are of discrete-sequence nature and of this size and diversity, as shown by Correia et al.~\cite{correia_dataset_2024-1}.
To adapt the PATH data set for active learning, the unlabelled data subset is split such that the time series within it represent a day's worth of measurements, two days worth of measurements, etc. until the entire subset is depleted, which results in $34$ increasingly larger splits.
Therefore, for the day $1$ split, the sum of the durations of each sequence within the split is roughly equal to $24$ hours, for the day 2 split the sum is roughly equal to $48$ hours, and so forth.
This mimics the gradual collection of data with time, like in the real world.
To maintain reasonable computational complexity, it is not feasible to perform active learning every day, therefore a querying round of $B$ samples is done after day $1$, day $7$, day $14$, day $21$, and day $28$.
On each of the days considered, a new model is also trained, such that the new threshold obtained using active learning is applied to the newly trained model.
For model training, the validation subset size is set as 20\% of the unlabelled subset, as is consistent with Correia et al.~\cite{correia_dataset_2024-1}.

To provide context to the results obtained, two baselines will also be provided to the results.
The first represents a scenario where the unsupervised threshold $\tau_\text{us}$ is used after each model training, whereas the second represents a hypothetical scenario where the best threshold $\tau_\text{best}$ is known.
It is obtained by grid-searching through the test subset and choosing the threshold yielding the highest $F_1$ score.
Neither of the scenarios involve active learning, i.e.\ no samples are sent to the oracle for labelling.
In addition to the results obtained using DQS, those for three other query strategies are presented.
The first, named the random-based query strategy (RQS), represents a scenario where, rather than intelligently querying an oracle, the samples to be labelled are chosen at random, as shown in Algorithm~\ref{alg:random_query_mech}.
\begin{algorithm}[h!]
\caption{Random-based query strategy at an arbitrary round $q$} \label{alg:random_query_mech}
\begin{algorithmic}[1]
\Require $\text{Candidate Score Set  } \mathcal{C}^{\text{as}}, \text{Query Score Set  } \mathcal{Q}^{\text{as}}$ 
\For{$b=1 \to B$}
    \State $r \sim \mathcal{U}(1, \vert \mathcal{C}^\text{as} \vert)$ \Comment{Random index is sampled}
    \State $\mathcal{Q}^{\text{as}} \gets \mathcal{C}^\text{as}[r]$
\EndFor
\State \textbf{return} $\text{Query Score Set  } \mathcal{Q}^{\text{as}}$
\end{algorithmic}
\end{algorithm}

The second, named the top-based query strategy (TQS)~\cite{tang_deep_2020, wang_practical_2020, bodor_little_2022, wang_active-mtsad_2022, huang_semi-supervised_2022, yu_amad_2024}, represents a scenario where the $B$ samples with the highest anomaly score are queried, as shown in Algorithm~\ref{alg:top_query_mech}.
Some literature~\cite{ning_deep_2022, russo_active_2020, zhou_boundary-driven_2024} also refers to this query strategy as uncertainty-based, as they interpret a high anomaly score as high uncertainty.
\begin{algorithm}[h!]
\caption{Top-based query strategy at an arbitrary round $q$} \label{alg:top_query_mech}
\begin{algorithmic}[1]
\Require $\text{Candidate Score Set  } \mathcal{C}^{\text{as}}, \text{Query Score Set  } \mathcal{Q}^{\text{as}}$ 
\For{$b=1 \to B$}
    \State $r = \arg\max(\mathcal{C}^{\text{as}})$ \Comment{Index of maximum anomaly score}
    \State $\mathcal{Q}^{\text{as}} \gets \mathcal{C}^\text{as}[r]$
\EndFor
\State \textbf{return} $\text{Query Score Set  } \mathcal{Q}^{\text{as}}$
\end{algorithmic}
\end{algorithm}

The third, named the uncertainty-based query strategy (UQS), represents a scenario where the $B$ samples closest to the threshold are queried, as shown in Algorithm~\ref{alg:uncertain_query_mech}.
This strategy needs to be initialised with a threshold, however, in literature, this is not always defined~\cite{wang_practical_2020, bodor_little_2022, wang_active-mtsad_2022} or it relies on labelled data~\cite{yu_amad_2024}.
\begin{algorithm}[h!]
\caption{
Uncertainty-based query strategy at an arbitrary round $q$
}
\label{alg:uncertain_query_mech}
\begin{algorithmic}[1]
\Require $\text{Candidate Score Set  } \mathcal{C}^{\text{as}}, \text{Query Score Set  } \mathcal{Q}^{\text{as}}$ 
\If{$\vert\mathcal{Q}^{\text{as}}\vert=0$}
    \State $\bar{\mathcal{C}}^{\text{as}}_n = \frac{1}{T_n}\sum^{T_n}_{t=1}\mathbf{s}_{n}$
    \State $\tau = \frac{1}{N}\sum^N_{n=1} (\bar{\mathcal{C}}^{\text{as}}_n)$
\EndIf
\For{$b=1 \to B$}
    \State $r = \arg\min(\text{abs}(\mathcal{C}^\text{as}-\tau))$ \Comment{Index of nearest anomaly score to $\tau$}
    \State $\mathcal{Q}^{\text{as}} \gets \mathcal{C}^\text{as}[r]$
\EndFor
\State \textbf{return} $\text{Query Score Set  } \mathcal{Q}^{\text{as}}$
\end{algorithmic}
\end{algorithm}

Like in Correia et al.~\cite{correia_dataset_2024-1}, the online evaluation metrics are obtained for each fold, though unlike in that paper, the modelling process isn't the focus.
The query and feedback strategies occur \emph{after} model training and highly depend on the seed used due to the random operations in the initialisation of the query strategies and mislabelling mechanism; therefore, we collect the results for each fold obtained using seeds 1 to 3, after which the average for each evaluation metric is provided.

The active learning-based approaches tested in this paper always use the most recent model available and assume that the oracle is queried once every day at the end of the day. 
For each split, labels for queried sequences in earlier splits are also considered.

The framework used for model training is TensorFlow 2.15.1 and TensorFlow Probability 0.23 on Python 3.10 on a workstation running Ubuntu 22.04 LTS, equipped with two Nvidia RTX A6000 GPUs.
TeVAE is trained as specified in~\cite{correia_tevae_2024, correia_dataset_2024-1}, except with varying training subset sizes.
Further information on library versions used can be found in the \emph{requirements.txt} file in the repository under \href{https://github.com/lcs-crr/DQS}{\texttt{github.com/lcs-crr/DQS}}.
    
\subsection{Results}
First, we present a selection of baseline results in Table~\ref{tab:results_baseline}.
The table shows the results obtained using the unsupervised threshold $\tau_\text{us}$, i.e.\ the threshold obtained by taking the maximum value anomaly score observed in the validation subset, and the results obtained hypothetical best threshold $\tau_\text{best}$, i.e.\ the threshold that maximises the $F_1$ score.
The latter is referred to as \emph{hypothetical} since it is obtained using the test subset, which in the real-world is not available, making this threshold unobservable.
\begin{table}[h!]
\centering
\caption{
$F_1$ scores for each of the splits using the unsupervised threshold $\tau_\text{us}$ and the hypothetical best threshold $\tau_\text{best}$.
}
\label{tab:results_baseline}
\resizebox{\textwidth}{!}{%
\begin{tabular}{lccccc}
      & 1 Day           & 7 Days          & 14 Days         & 21 Days         & 28 Days          \\ \hline \hline 
$\tau_\text{us}$  & $0.16 \pm 0.03$ & $0.16 \pm 0.01$ & $0.12 \pm 0.10$ & $0.06 \pm 0.04$ & $0.04 \pm 0.02$  \\
$\tau_\text{best}$          & $0.20 \pm 0.05$ & $0.38 \pm 0.09$ & $0.45 \pm 0.10$ & $0.58 \pm 0.13$ & $0.68 \pm 0.10$  \\ \hline 
\end{tabular}%
}
\end{table}

As expected, the results using the hypothetical best threshold $\tau_\text{best}$ are higher across the board.
There is a downward trend in the $F_1$ score obtained with the unsupervised threshold $\tau_\text{us}$ as time goes on and the validation subset grows.
This is because, with a larger validation subset and therefore a higher anomaly count within it, the probability of a very high anomaly score grows, leading to a higher-than-ideal threshold setting.
\emph{The goal of any query strategy is to outperform the unsupervised threshold $\tau_\text{us}$, while coming as close to the hypothetical best as possible.}

\begin{figure}[h!]
    \centering
    \def\svgwidth{\textwidth}
    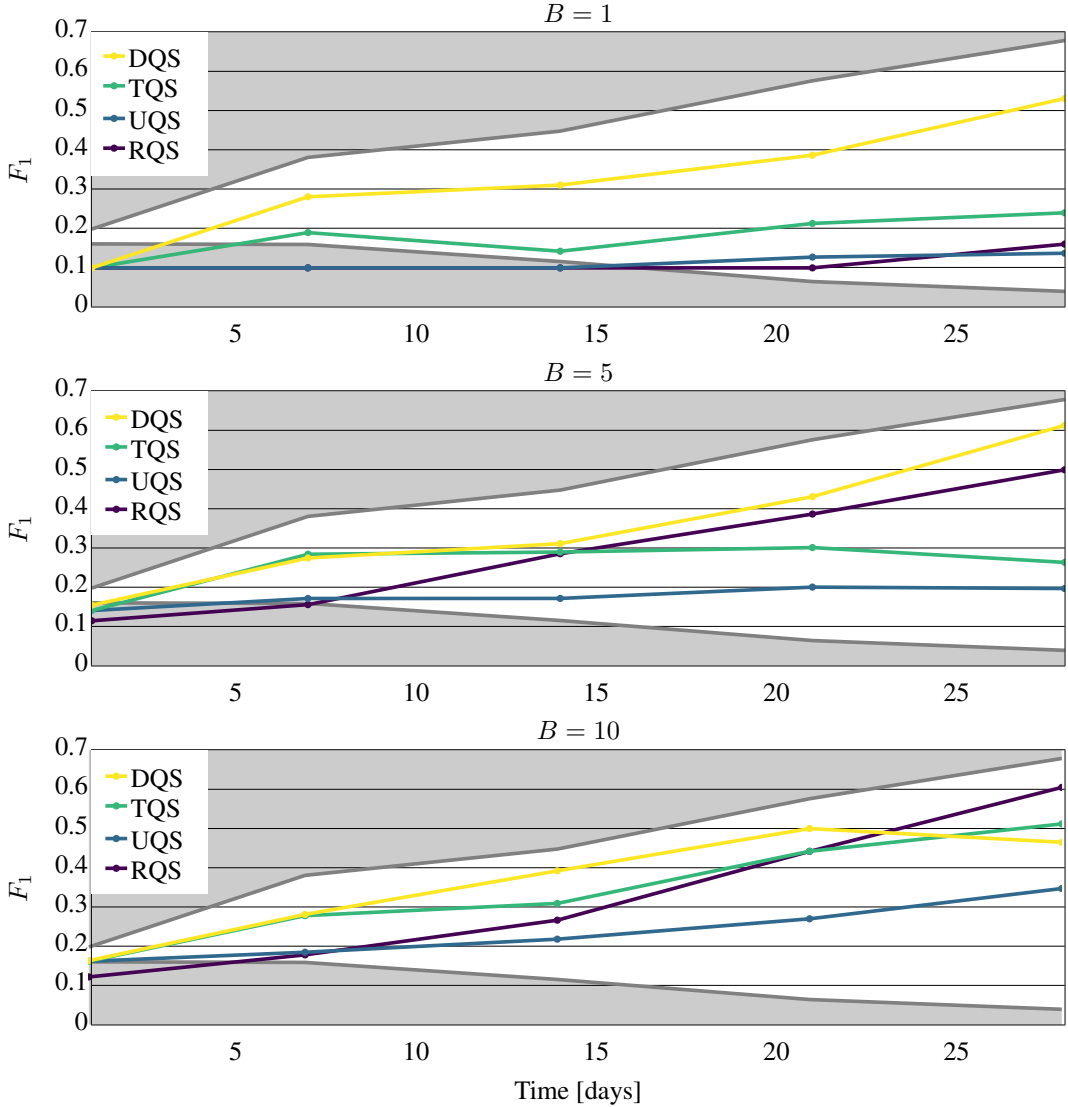
    \caption{
    $F_1$ scores for each query strategy plotted as a function of time. 
    The results in tabular form are also shown in Table~\ref{tab:results_budgets}.
    The grey upper bound represents the results using the hypothetical best threshold $\tau_\text{best}$, and the grey lower bound represents the results using the unsupervised threshold $\tau_\text{us}$.
    }
    \label{fig:results_budgets}
\end{figure}

The results for different query strategies (random-based, top-based, uncer\-tainty-based, dissimilarity-based) and different query budgets ($B=1$, $B=5$, $B=10$) are presented in Table~\ref{tab:results_budgets} and, to further aid interpretation, as a function of time in Figure~\ref{fig:results_budgets}.
\begin{table}[h!]
\centering
\caption{
$F_1$ scores for each of the splits for different query budgets using different query strategies and the DQS (ours).
The results obtained using the unsupervised threshold $\tau_\text{us}$ and the hypothetical best threshold $\tau_\text{best}$ are also presented for reference.
The best results for the respective split at a given budget size $B$ is shown in \textbf{bold}. 
The values shown consist of the mean $\pm$ standard deviation over the three folds, each on the three different seeds.
A plot of the $F_1$ score as a function of time is shown in Figure~\ref{fig:results_budgets}.
}
\label{tab:results_budgets}
\resizebox{\textwidth}{!}{%
\begin{tabular}{lcccccc}
                 & $B$ & 1 Day           & 7 Days          & 14 Days         & 21 Days         & 28 Days        \\ \hline\hline
$\tau_\text{us}$ & - & $0.16 \pm 0.03$ & $0.16 \pm 0.01$ & $0.12 \pm 0.10$ & $0.06 \pm 0.04$ & $0.04 \pm 0.02$  \\
$\tau_\text{us}$ & - & $0.20 \pm 0.05$ & $0.38 \pm 0.09$ & $0.45 \pm 0.10$ & $0.58 \pm 0.13$ & $0.68 \pm 0.10$  \\ \hline 
RQS & 1          & $\mathbf{0.10 \pm 0.01}$ & $0.10 \pm 0.01$ & $0.10 \pm 0.01$ & $0.10 \pm 0.01$ & $0.16 \pm 0.16$ \\
TQS & 1          & $\mathbf{0.10 \pm 0.01}$ & $0.19 \pm 0.13$ & $0.14 \pm 0.05$ & $0.21 \pm 0.14$ & $0.24 \pm 0.17$ \\
UQS & 1          & $\mathbf{0.10 \pm 0.01}$ & $0.10 \pm 0.01$ & $0.10 \pm 0.01$ & $0.13 \pm 0.04$ & $0.14 \pm 0.06$ \\
DQS & 1          & $\mathbf{0.10 \pm 0.01}$ & $\mathbf{0.28 \pm 0.11}$ & $\mathbf{0.31 \pm 0.14}$ & $\mathbf{0.39 \pm 0.18}$ & $\mathbf{0.53 \pm 0.19}$ \\ \hline
RQS & 5          & $0.11 \pm 0.02$ & $0.16 \pm 0.06$ & $0.29 \pm 0.18$ & $0.39 \pm 0.23$ & $0.50 \pm 0.25$ \\
TQS & 5          & $0.14 \pm 0.06$ & $\mathbf{0.28 \pm 0.06}$ & $0.29 \pm 0.02$ & $0.30 \pm 0.06$ & $0.26 \pm 0.12$ \\
UQS & 5          & $0.14 \pm 0.06$ & $0.17 \pm 0.11$ & $0.17 \pm 0.11$ & $0.20 \pm 0.15$ & $0.20 \pm 0.14$ \\
DQS & 5          & $\mathbf{0.15 \pm 0.05}$ & $0.27 \pm 0.07$ & $\mathbf{0.31 \pm 0.15}$ & $\mathbf{0.43 \pm 0.19}$ & $\mathbf{0.61 \pm 0.13}$ \\ \hline
RQS & 10         & $0.12 \pm 0.03$ & $0.18 \pm 0.06$ & $0.27 \pm 0.17$ & $0.44 \pm 0.22$ & $\mathbf{0.60 \pm 0.15}$ \\
TQS & 10         & $\mathbf{0.16 \pm 0.05}$ & $\mathbf{0.28 \pm 0.07}$ & $0.31 \pm 0.12$ & $0.44 \pm 0.06$ & $0.51 \pm 0.14$ \\
UQS & 10         & $\mathbf{0.16 \pm 0.05}$ & $0.18 \pm 0.07$ & $0.22 \pm 0.14$ & $0.27 \pm 0.12$ & $0.35 \pm 0.18$ \\
DQS & 10         & $\mathbf{0.16 \pm 0.05}$ & $\mathbf{0.28 \pm 0.06}$ & $\mathbf{0.39 \pm 0.10}$ & $\mathbf{0.50 \pm 0.12}$ & $0.46 \pm 0.23$ \\ \hline
\end{tabular}%
}
\end{table}

The first observation that can be made from Table~\ref{tab:results_budgets} and Figure~\ref{fig:results_budgets} is that, at the smallest split of 1 day for all budget sizes, all query strategies perform worse or at most comparably to the scenario using the unsupervised threshold $\tau_\text{us}$.
This can be attributed to the poor model quality at the small training subset size, indicated by the low hypothetical best results.
\emph{As the training subset grows, the difference between the two baselines grows, which is where active learning can have the most impact.}
At small budget sizes, i.e.\ at $B<10$, the DQS outperforms the other query strategies across the board. 
At $B=10$ the DQS outperforms or is at most matched by the other query strategies except for the last split of $28$ days, where the RQS performs the best.
Lastly, at most budget sizes and splits, the UQS falls short of the other query strategies.

\begin{figure}[h!]
    \centering
    \def\svgwidth{\textwidth}
    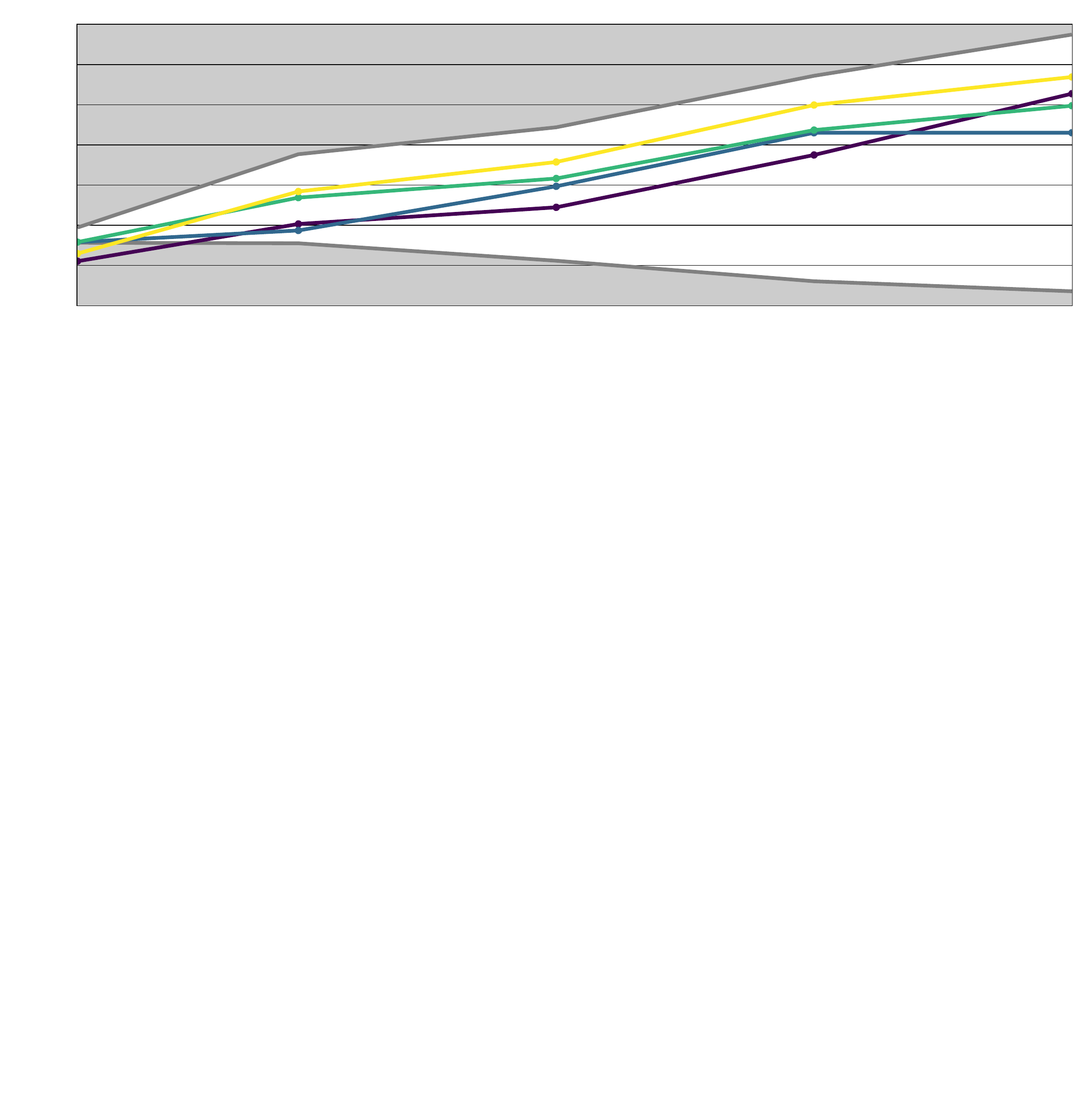
    \caption{
    $F_1$ scores for each query strategy plotted as a function of time.
    The results in tabular form are also shown in Table~\ref{tab:results_mislabelling}.
    The grey upper bound represents the results using the hypothetical best threshold $\tau_\text{best}$, and the grey lower bound represents the results using the unsupervised threshold $\tau_\text{us}$.
    }
    \label{fig:results_mislabelling}
\end{figure}

The results for different query strategies (random-based, top-based, uncer\-tainty-based, dissimilarity-based) and different mislabelling probabilities ($p_m=0.1$, $p_m=0.2$, $p_m=0.3$) are presented in Table~\ref{tab:results_mislabelling} and, to further aid interpretation, as a function of time in Figure~\ref{fig:results_mislabelling}.
\begin{table}[h!]
\centering
\caption{
$F_1$ scores for each of the splits for a query budget of $B=10$ and different mislabelling probabilities using different query strategies and the DQS (ours).
The results obtained using the unsupervised threshold $\tau_\text{us}$ and the hypothetical best threshold $\tau_\text{best}$ are also presented for reference.
The best results for the respective split at a given mislabelling probability $p_m$ is shown in \textbf{bold}.
The values shown consist of the mean $\pm$ standard deviation over the three folds, each on the three different seeds.
A plot of the $F_1$ score as a function of time is shown in Figure~\ref{fig:results_mislabelling}.
}
\label{tab:results_mislabelling}
\resizebox{\textwidth}{!}{%
\begin{tabular}{lcccccc}
                   & $p_m$ & 1 Day           & 7 Days          & 14 Days         & 21 Days         & 28 Days \\ \hline \hline 
$\tau_\text{us}$   & - & $0.16 \pm 0.03$ & $0.16 \pm 0.01$ & $0.12 \pm 0.10$ & $0.06 \pm 0.04$ & $0.04 \pm 0.02$  \\
$\tau_\text{best}$ & - & $0.20 \pm 0.05$ & $0.38 \pm 0.09$ & $0.45 \pm 0.10$ & $0.58 \pm 0.13$ & $0.68 \pm 0.10$  \\ \hline 
RQS & 0            & $0.12 \pm 0.03$ & $0.18 \pm 0.06$ & $0.27 \pm 0.17$ & $0.44 \pm 0.22$ & $\mathbf{0.60 \pm 0.15}$ \\
TQS & 0     & $\mathbf{0.16 \pm 0.05}$ & $\mathbf{0.28 \pm 0.07}$ & $0.31 \pm 0.12$ & $0.44 \pm 0.06$ & $0.51 \pm 0.14$ \\
UQS & 0     & $\mathbf{0.16 \pm 0.05}$ & $0.18 \pm 0.07$ & $0.22 \pm 0.14$ & $0.27 \pm 0.12$ & $0.35 \pm 0.18$ \\
DQS & 0     & $\mathbf{0.16 \pm 0.05}$ & $\mathbf{0.28 \pm 0.06}$ & $\mathbf{0.39 \pm 0.10}$ & $\mathbf{0.50 \pm 0.12}$ & $0.46 \pm 0.23$ \\ \hline
RQS & 0.1   & $0.11 \pm 0.02$ & $0.21 \pm 0.06$ & $0.25 \pm 0.15$ & $0.38 \pm 0.17$ & $0.53 \pm 0.21$ \\
TQS & 0.1   & $\mathbf{0.16 \pm 0.05}$ & $0.27 \pm 0.07$ & $0.32 \pm 0.06$ & $0.44 \pm 0.19$ & $0.50 \pm 0.13$ \\
UQS & 0.1   & $\mathbf{0.16 \pm 0.05}$ & $0.19 \pm 0.09$ & $0.30 \pm 0.12$ & $0.43 \pm 0.22$ & $0.43 \pm 0.11$ \\
DQS & 0.1   & $0.13 \pm 0.05$ & $\mathbf{0.29 \pm 0.06}$ & $\mathbf{0.36 \pm 0.13}$ & $\mathbf{0.50 \pm 0.15}$ & $\mathbf{0.57 \pm 0.17}$ \\ \hline
RQS & 0.2   & $0.12 \pm 0.02$ & $0.17 \pm 0.06$ & $0.20 \pm 0.14$ & $0.20 \pm 0.10$ & $0.43 \pm 0.17$ \\
TQS & 0.2   & $\mathbf{0.16 \pm 0.05}$ & $\mathbf{0.33 \pm 0.08}$ & $\mathbf{0.33 \pm 0.06}$ & $0.43 \pm 0.05$ & $0.51 \pm 0.15$ \\
UQS & 0.2   & $\mathbf{0.16 \pm 0.05}$ & $0.23 \pm 0.11$ & $0.29 \pm 0.12$ & $0.44 \pm 0.24$ & $0.46 \pm 0.16$ \\
DQS & 0.2   & $0.11 \pm 0.04$ & $0.23 \pm 0.10$ & $\mathbf{0.33 \pm 0.13}$ & $\mathbf{0.46 \pm 0.17}$ & $\mathbf{0.54 \pm 0.17}$ \\ \hline
RQS & 0.3   & $0.11 \pm 0.02$ & $0.15 \pm 0.06$ & $0.20 \pm 0.16$ & $0.15 \pm 0.05$ & $0.21 \pm 0.12$ \\
TQS & 0.3   & $\mathbf{0.14 \pm 0.06}$ & $\mathbf{0.35 \pm 0.06}$ & $0.28 \pm 0.12$ & $\mathbf{0.52 \pm 0.15}$ & $0.47 \pm 0.14$ \\
UQS & 0.3   & $0.14 \pm 0.06$ & $0.26 \pm 0.12$ & $\mathbf{0.31 \pm 0.12}$ & $0.37 \pm 0.26$ & $0.40 \pm 0.15$ \\
DQS & 0.3   & $0.11 \pm 0.04$ & $0.23 \pm 0.11$ & $0.30 \pm 0.17$ & $0.33 \pm 0.21$ & $\mathbf{0.55 \pm 0.15}$ \\ \hline
\end{tabular}%
}
\end{table}

Unsurprisingly, the above-made observations regarding the low $F_1$ scores at the smallest split, can also be made when investigating the impact of mislabelling on detection performance.
Aside from the results at the smallest split, all query strategies perform comparably at $p_m=0.1$, with DQS performing slightly better.
This holds mostly true for $p_m=0.2$ and $p_m=0.3$ as there is no clear winner for all split sizes.
What is clear, is that at the two higher mislabelling probabilities, the RQS performs worst.
Overall, the results scatter more between query strategies as the mislabelling probability increases.
Interestingly, some results improve despite higher mislabelling probabilities.
This can happen when the mislabelled samples do not change the position of the threshold, for example, when an equal number of samples are mislabelled on either side of the threshold.

\section{Conclusion and Outlook} \label{sec:conclusion}
In this work, we combine active learning with an existing unsupervised anomaly detection approach by selectively querying the label of time series and using the obtained labels to search for a more suitable threshold.
To this end, we propose a novel query strategy, the dissimilarity-based query strategy (DQS), which attempts to maximise the diversity of queried samples, by inspecting the similarity between resulting anomaly scores using dynamic time warping, which allows anomaly scores to be of variable-length.
We compare the detection performance resulting from DQS with a number of other query strategies and also investigate the impact of mislabelling, which had been unexplored in literature until now.

The results show that, there is no query strategy that outperforms all others in every category.
While DQS performs best in low-budget cases and remains competitive at higher ones, TQS demonstrates higher robustness when faced with mislabelling. 
In the real world, it therefore depends on the expertise of the oracle and how many samples they are willing to label.
In any case, except for very small splits, all querying strategies outperform the unsupervised threshold $\tau_\text{us}$ despite the presence of mislabelling, and hence it is advisable to use an active learning-based threshold if the possibility to query an oracle exists.

Future work should focus on standardising the benchmarking procedure of active learning methods in time series anomaly detection and should include investigation on the impact of mislabelling, proposed in this work. 
Furthermore, creating a query strategy that not only performs well with small query budgets but is also robust in the case of mislabelling should be further researched.

\printbibliography 

\end{document}